\documentclass[letterpaper]{article} 
\usepackage{aaai25}  
\usepackage{times}  
\usepackage{helvet}  
\usepackage{courier}  
\usepackage[hyphens]{url}  
\usepackage{graphicx} 
\usepackage{natbib}  
\usepackage{caption} 
\usepackage{algorithm}
\usepackage{algorithmic}

\usepackage{amsmath}
\usepackage{amssymb}
\usepackage{makecell}

\usepackage{newfloat}
\usepackage{listings}
\DeclareCaptionStyle{ruled}{labelfont=normalfont,labelsep=colon,strut=off} 
\floatstyle{ruled}
\newfloat{listing}{tb}{lst}{}
\floatname{listing}{Listing}
\usepackage{booktabs}

\title{VI3DRM:Towards meticulous 3D Reconstruction from Sparse Views \\ via Photo-Realistic Novel View Synthesis}
\author{
    Hao Chen$^1$\thanks{Equal contribution.},
    Jiafu Wu$^2$\footnotemark[1],
    Ying Jin$^1$,
    Jinlong Peng$^2$,
    Xiaofeng Mao$^1$,\\
    Mingmin Chi$^1$\thanks{Corresponding author.},
    Mufeng Yao$^1$,
    Bo Peng$^3$,
    Jian Li$^2$,
    Yun Cao$^2$,
}
\affiliations{
    \textsuperscript{\rm 1}Fudan University, 
    \textsuperscript{\rm 2}Tencent Youtu Lab,
    \textsuperscript{\rm 3}Shanghai Ocean University\\
    \texttt{haochen22@m.fudan.edu.cn; jiafwu@tencent.com}

}

\usepackage{bibentry}

\begin{document}
\maketitle



\begin{abstract}
Recently, methods like Zero-1-2-3 have focused on single-view based 3D reconstruction and have achieved remarkable success. However, their predictions for unseen areas heavily rely on the inductive bias of large-scale pretrained diffusion models. Although subsequent work, such as DreamComposer, attempts to make predictions more controllable by incorporating additional views, the results remain unrealistic due to feature entanglement in the vanilla latent space, including factors such as lighting, material, and structure. To address these issues, we introduce the \textbf{V}isual \textbf{I}sotropy \textbf{3D} \textbf{R}econstruction \textbf{M}odel (VI3DRM), a diffusion-based sparse views 3D reconstruction model that operates within an ID-consistent and perspective-disentangled 3D latent space. By facilitating the disentanglement of semantic information, color, material properties and lighting, VI3DRM is capable of generating highly realistic images that are indistinguishable from real photographs. By leveraging both real and synthesized images, our approach enables the accurate construction of pointmaps, ultimately producing finely textured meshes or point clouds. On the NVS task, tested on the GSO dataset, VI3DRM significantly outperforms state-of-the-art method DreamComposer, achieving a PSNR of \textbf{38.61 ($\uparrow$ 42\%)}, an SSIM of \textbf{0.929 ($\uparrow$ 2\%)}, and an LPIPS of \textbf{0.027 ($\downarrow$ 63\%)}. Code will be made available upon publication.

\end{abstract}



%


\section{1.Introduction}


Image-based 3D reconstruction has long been a significant and challenging task in the field of Computer Vision, with critical applications in areas such as 3D asset generation, Augmented Reality (AR), and Virtual Reality (VR). Traditional multi-view stereo (MVS) methods often grapple with the trade-off between reconstruction quality and practicality, especially in terms of the number of images required. This trade-off remains a major hurdle for real-world applications. Consequently, the ability to generate high-quality 3D content from a limited number of viewpoints poses a more formidable challenge and holds greater practical value.

\begin{figure}[t]
\centering
\includegraphics[width=1\columnwidth]{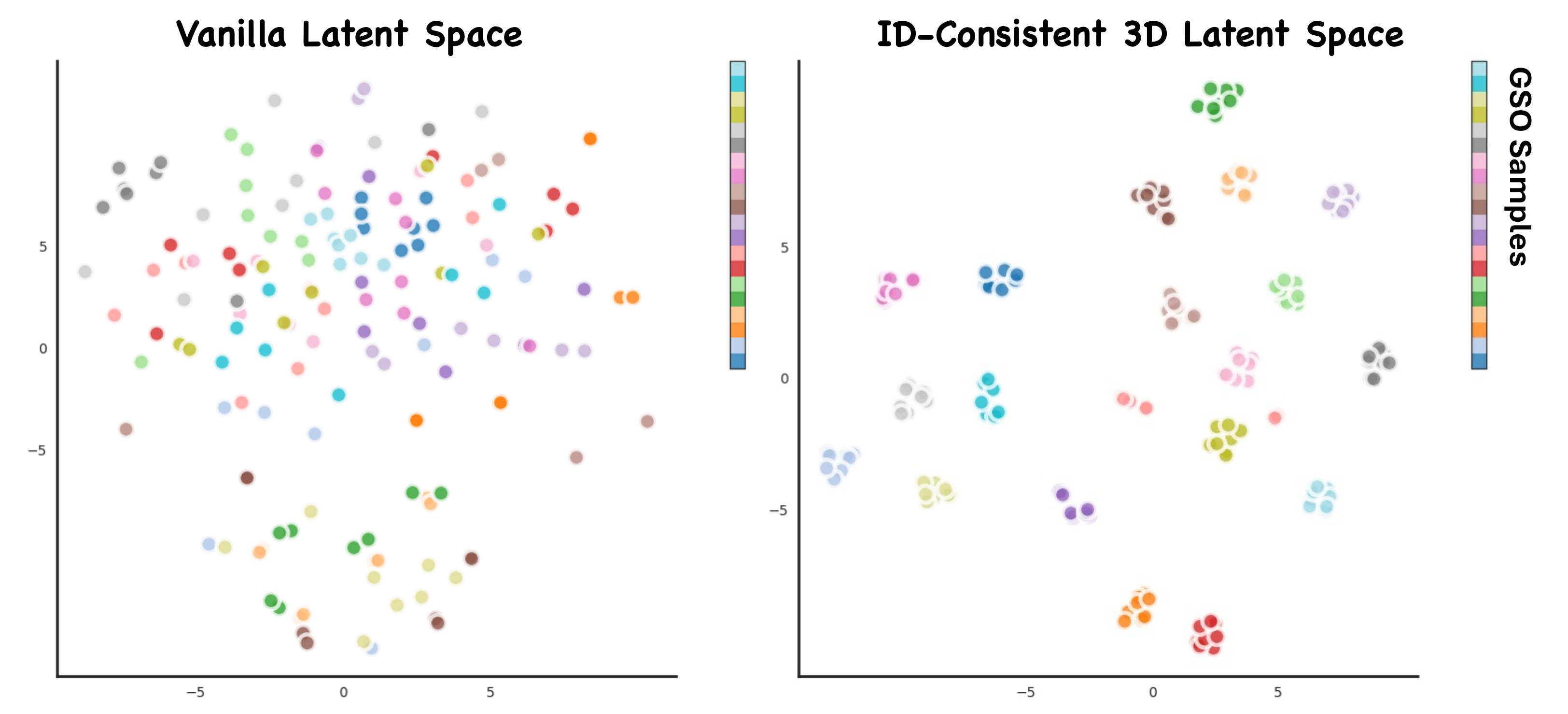} 
\caption{Latent Visualization. Twenty randomly sampled GSO objects are encoded in both the Vanilla Latent Space (left) and our ID-Consistent Latent Space (right). The original feature visualization (left) is scattered and disordered, whereas ours tightly clusters different views of the same object in the latent space.}
\label{fig:tsne}
\end{figure}

\begin{figure*}[t]
\centering
\includegraphics[width=1\textwidth]{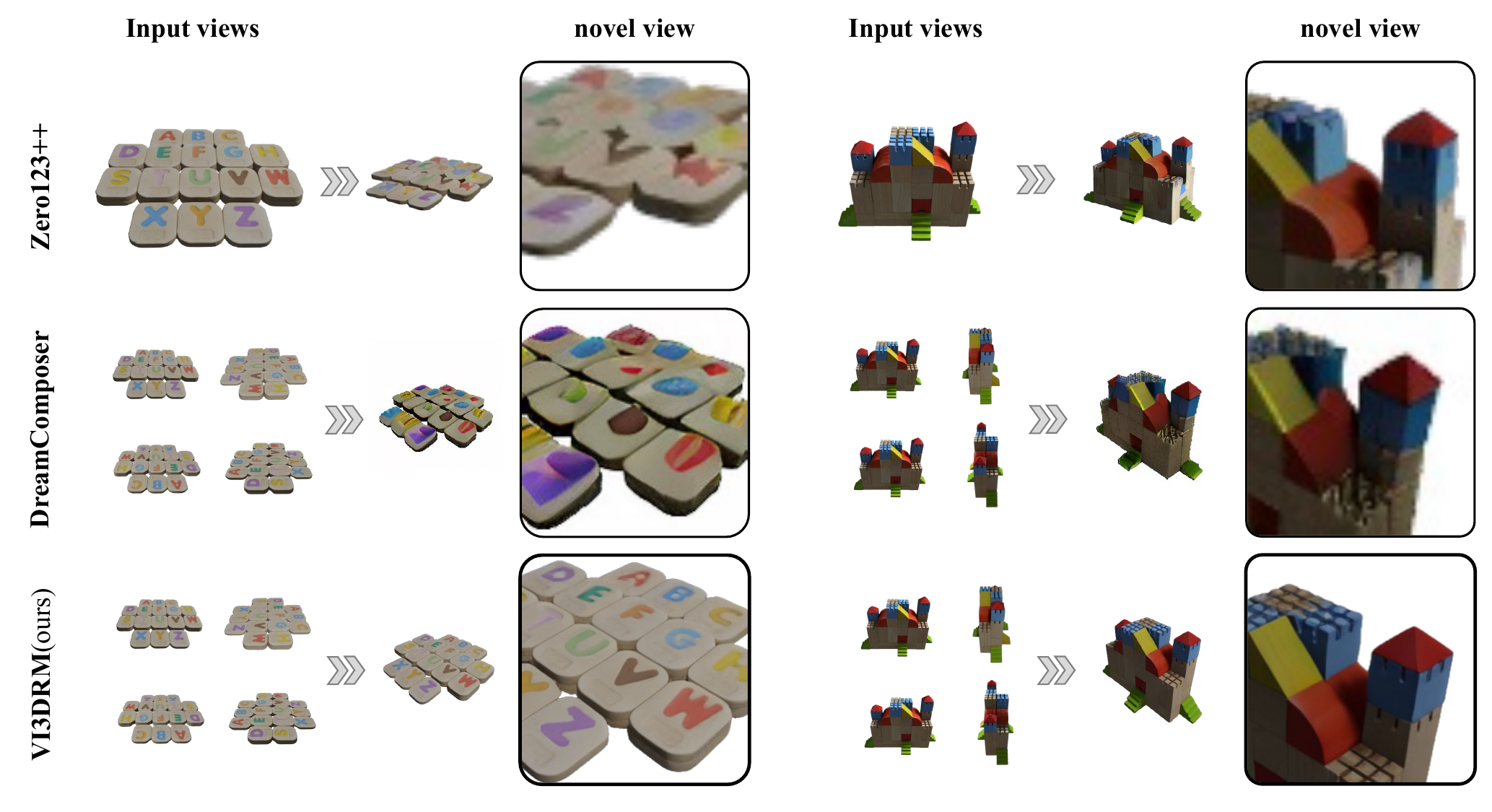} 
\caption{Zero-shot NVS on GSO dataset, our method outperforms previous approaches by a large margin in terms of both texture and structural accuracy}
\label{fig:GSO}
\end{figure*}


Recently, the successful application of 2D conditional diffusion models in Novel View Synthesis (NVS) has paved the way for reducing the dependence on acquisition viewpoints by leveraging large-scale 2D prior knowledge, such as stable diffusion \cite{rombach2022high}. Studies like Zero-1-2-3 \cite{liu2023zero}, Zero123++ \cite{shi2023zero123++}, One-2-3-45 \cite{liu2024one}, One-2-3-45++\cite{liu2024one2345++} employ a 2D conditional diffusion model to generate consistent novel views, which are then lifted into 3D representations through MVS methods, whether optimize-based or feed-forward. Despite the significant advancements these works have achieved, starting from a single viewpoint fails to capture comprehensive structural information, thereby limiting their applicability to specific use cases. Furthermore, previous diffusion-based NVS methods have encountered difficulties in generating new perspectives with accurate structural consistency, scene lighting consistency, and detailed textures. The synthesized images produced by these models often deviate significantly from the actual viewpoints, resulting in substantial errors that negatively affect the quality of subsequent 3D reconstructions. Although methods like One-2-3-45 \cite{liu2024one} have attempted to bridge the gap between synthesized and real images, the results remain less than optimal.

To enhance the controllability of novel view synthesis, DreamComposer \cite{yang2024dreamcomposer} attempts to incorporate multi-view conditions into the diffusion models using a ControlNet-like structure \cite{zhang2023adding}. Despite offering improved controllability compared to Zero-1-2-3, DreamComposer still struggles to generate realistic images. This lack of realism is evident in inaccurate object structures, loss of surface texture details, and inconsistent environmental lighting. As shown in Figure \ref{fig:tsne}, The reason behind this is that the traditional latent space lacks the capability to identify object identities, making it difficult for the diffusion model to capture the complex mapping relationships across different views.

To address these issues, we introduce the Visual Isotropy 3D Reconstruction Model (VI3DRM), a model that fundamentally utilizes diffusion-based Novel View Synthesis (NVS) at its core. This model facilitates the mapping from known to novel views within a 3D latent space that is both identity-consistent and perspective-disentangled. Our space excels at modeling perspective-independent attributes of an object, including semantic information, color, material properties, and lighting conditions, facilitating the preservation of identity in Novel View Synthesis. Furthermore, we employ a pseudo cross-attention approach to facilitate the exchange of information between perspectives.

 As shown in Figure \ref{fig:GSO}, our model can generate novel views that are virtually indistinguishable from real photographs, outperforming previous models in terms of both texture details and structural consistency. When evaluated on the NVS task using the GSO dataset \cite{downs2022google}, our VI3DRM shows significant improvements over the state-of-the-art algorithm DreamComposer \cite{yang2024dreamcomposer}. Specifically, the Peak Signal-to-Noise Ratio (PSNR) has increased from 27.21 to 38.61, the Structural Similarity Index Measure (SSIM) has improved from 0.906 to 0.929, and the Learned Perceptual Image Patch Similarity (LPIPS) has decreased from 0.073 to 0.027. Leveraging realistic synthesis images, we utilize dust3r \cite{dust3rpose} to produce high-quality textured meshes within 60 seconds.

\section{2.Related Work}

\subsection{2.1 Single-View based 3D Reconstruction}

Recent advancements in 3D reconstruction from a single view have achieved remarkable results. These methods typically involve two primary steps: initially, a single-view based diffusion model is utilized to predict novel views, which are subsequently processed using multi-view stereo (MVS) techniques to generate a mesh or other 3D representation. Zero123 \cite{liu2023zero} was among the pioneers to exploit the rich geometric information embedded in the pre-trained Stable Diffusion model \cite{rombach2022high}, using a 2D conditional diffusion model to generate novel views. Building upon this, Zero123++ \cite{shi2023zero123++} integrated several techniques to enhance the consistency of the generated images. To improve the consistency of synthetic novel viewpoint images, SV3D \cite{voleti2024sv3d} harnesses the object consistency characteristics of robust stable video diffusion models. 

Despite the advancements made by these studies, their predictions heavily rely on the inductive bias in the diffusion model, such as stable diffusion or stable video diffusion. Moreover, the generated results still lag behind real images in terms of accuracy, structural consistency, lighting, and texture. To mitigate the gap caused by the inaccuracies in the generated images, additional Mixed Training is required to reduce the reconstruction error caused by the imperfect initial stage in One-2-3-45\cite{liu2024one}.

\subsection{2.2 Sparse-View based 3D Reconstruction}

Due to the scarcity of large-scale 3D datasets, a prevalent approach in few-views 3D reconstruction today leverages the prior information embedded in 2D diffusion models that have been pre-trained on extensive 2D datasets. Notable efforts, such as DreamField\cite{jain2022zero} and DreamFusion\cite{poole2022dreamfusion}, employ CLIP\cite{radford2021learning} semantic feature similarity loss or SDS loss to guide the optimization of neural radiance fields. However, these optimization-based methods are time-intensive, often requiring several hours or days to optimize a single scene, and typically yield results that are overly smooth and lack fine details. An alternative strategy involves using a 2D conditional diffusion model to synthesize new viewpoint images, thereby providing additional supervisory information for two-stage image-to-3D algorithms.

To address the inherent limitations of single-view Novel View Synthesis (NVS), DreamComposer\cite{yang2024dreamcomposer} aims to enhance prediction controllability by incorporating the latent features from additional views. Although DreamComposer improves upon the controllability seen in Zero123\cite{liu2023zero} and SyncD\cite{liu2023syncdreamer}, the newly generated view images still fall short in terms of accuracy and detail.

In contrast to these incremental, plug-in-style enhancements, our approach constructs a latent space designed to preserve object information consistently across different viewpoints. We adopt a more direct and efficient method for fusing features from known viewpoints, significantly outperforming previous algorithms on multi-view based NVS tasks.

\subsection{2.3 MVS Methods}

Existing 3D reconstruction methods can generally be divided into two categories: optimization-based and feed-forward. The optimization-based approach has been widely studied in recent years following the emergence of NeRF \cite{mildenhall2021nerf} and 3D Gaussian \cite{kerbl20233d} methods. To extract 3D reconstruction from the neural field or 3d Gaussian, some studies attempt to optimize the surface through the design of regularization terms, like SuGaR\cite{guedon2024sugar}, 2DGS\cite{huang20242d}, while neus \cite{wang2021neus} combine signed distance function (SDF) with implicit Neural Radiance Fields. Despite their remarkable results, these techniques require numerous views for relatively accurate output and are extremely time-consuming, often taking minutes or hours per 3D shape.

Depth-map based methods decouple the complex MVS problem into relatively small problems of per-view depth map estimation in a feed-forward manner. The MVSNet\cite{yao2018mvsnet} along with its subsequent enhancements focus on depth map estimation from one reference image and a few source images at a time, offering greater flexibility. The recent open-source project Dust3r \cite{wang2024dust3r} converts the feature maps of other reference views to the current view through projection, and combines information from multiple views to predict accurate pointmaps. However, Dust3r struggles to get accurate depth estimates from view projections of sparse views, especially when these views are far apart. Through realistic and unbiased synthesized novel views, our model enables dust3r to exhibit greater robustness and construct more accurate pointmaps with just 4 views. Additionally, our NVS model can seamlessly enhance the performance of any other MVS models, making them more versatile and practical.


\section{3. Method}
\subsection{3.1 Multi-View 3D Reconstruction definition}
\begin{figure*}[ht]
\centering
\includegraphics[width=1\textwidth]{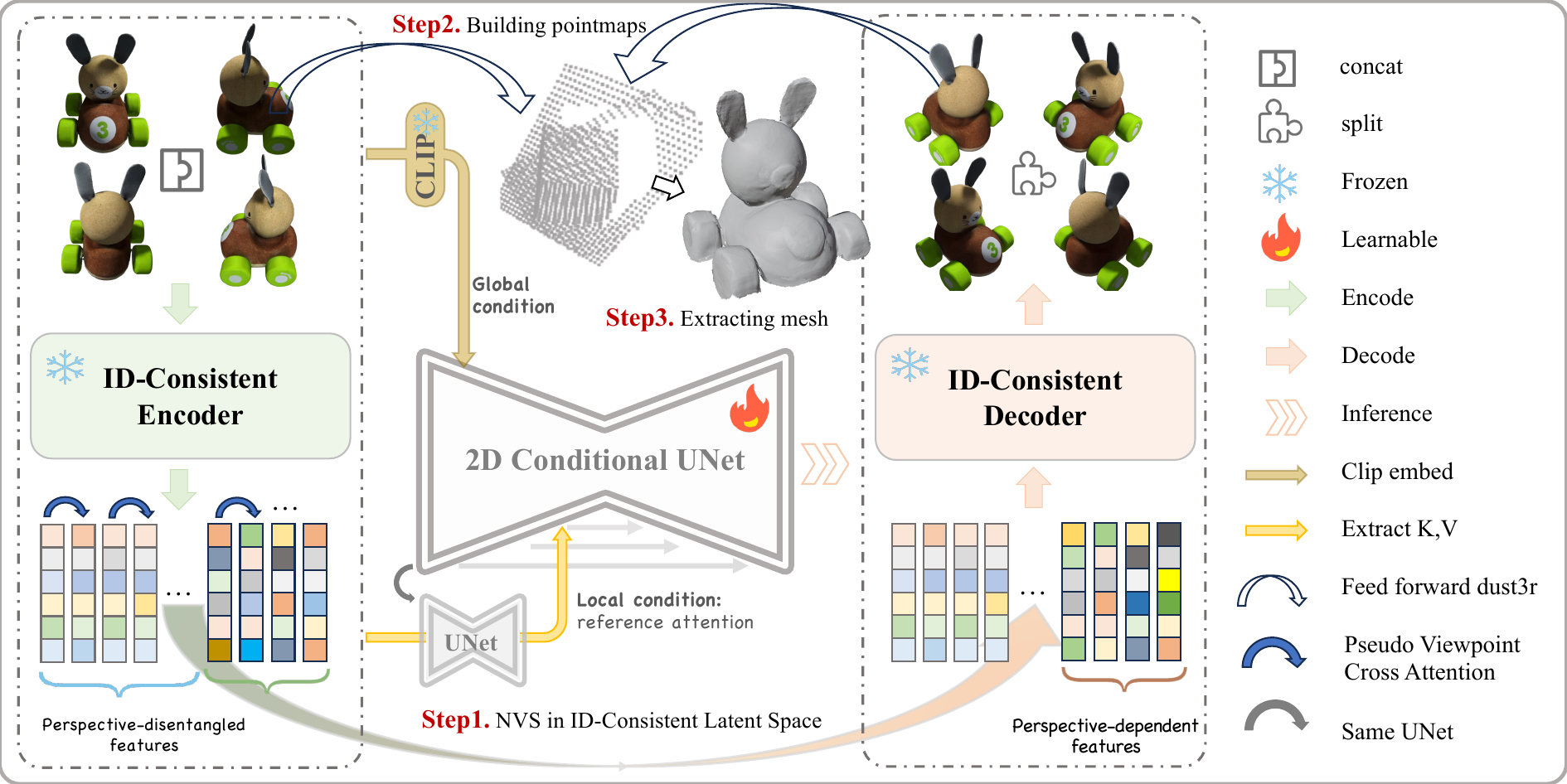} 
\caption{Our method involves three primary steps: \textbf{Step 1}: We encode four known view images into our ID-consistent Latent Space and extract semantic embeddings as global condition. These conditions guide the LDM model to generate four novel views. \textbf{Step 2}: We feed both the original and synthesized images into Dust3r to construct optimized pointmaps. \textbf{Step 3}: Meshes or point clouds are extracted from the pointmaps. }
\label{fig:method}
\end{figure*}

As depicted in Figure \ref{fig:method}, our approach processes RGB images from four distinct camera viewpoints of the same object. The goal is to produce a precise 3D representation that faithfully reconstructs every occluded region of the real object with both realism and consistency. A high-quality multi-view 3D mesh reconstruction should demonstrate fidelity in several critical aspects: continuity and isotropy of edges, clear delineation of occluded regions, and physical realism when viewed from arbitrary perspectives.

Instead of directly producing full 3D reconstructions from images captured from multiple camera viewpoints, our approach first establishes a consistent 3D representation within the latent space, conditioned on the initial set of multi-view images. Leveraging these 3D latent codes, our method then generates novel view synthesis via a 3D decoder. This decoder is further informed by the aligned representations and occluded detail features extracted by the 3D encoder from the input images.

\subsection{3.2  ID-Consistent and Perspective Disentangled 3D Latent Space }
Given a set of unlabeled four-viewpoints images of 3D objects, we aim to build a latent space for 3D geometry structure with high degrees of perspective disentangled and ID-consistent. The disentanglement facilitates effective generative modeling of semantic information, color, material properties, and lighting across large sets of image pairs, regardless of the subject identities. It also enables the manipulation of disentangled factors in the output, which is particularly advantageous for multi-view 3D reconstruction tasks. Existing methods frequently exhibit deficiencies in ID consistency, and detail representation from occluded viewpoints. On the other hand, comprehensive and consistent detail expressiveness ensures that the decoder can produce high-quality novel view synthesis with rich texture details, while the latent generator effectively captures the subtle occluded areas of the 3D object.

To achieve this, we decompose a canonical 3D object into a 3D texture detail code $\mathbf{z}^{\text {text }}$, an identity code $\mathbf{z}^{\text {id }}$, and a 3D holistic appearance volume $\mathbf{z}^{\text {app }}$. Each of them is extracted from a 3D render image by a comprehensive encoder. A single decoder $\mathcal{D}$ takes these latent variables as input and reconstructs the 3D image, where similar warping fields in the inverse direction are first applied to $\mathbf{z}^{a p p}$ to get the four-viewpoints appearance.

To learn the perspective disentangled latent space, the core idea is to construct image reconstruction loss by swapping latent variables between different images in 3D object different viewpoints. We observed that the original KL reconstruction loss demonstrates insufficient disentanglement between the 3D holistic appearance and 3D texture detail. Images rendered from different viewpoints of the same object naturally share similarities in semantic information, color, material, and lighting, which should result in close distances. Let \(\mathbf{I}_i\) and \(\mathbf{I}_j\) be two viewpoint images randomly rendered from the same 3D object. We extract their latent variables \(\hat{\mathbf{I}}_{i}=\mathcal{D}\left(\mathbf{v}_i^{\text{dir}}, \mathbf{z}_i^{\text{app}}, \mathbf{z}_i^{\text{id}}, \mathbf{z}_i^{\text{text}}\right)\) using the encoders, where \(\mathbf{v}_i^{\text{dir}}\) represents the camera viewpoint. If the texture detail information between \(\mathbf{I}_i\) and \(\mathbf{I}_j\) is swapped, i.e., \(\hat{\mathbf{I}}_{i, \mathbf{z}_j^{\text{text}}}=\mathcal{D}\left(\mathbf{v}_i^{\text{dir}}, \mathbf{z}_i^{\text{app}}, \mathbf{z}_i^{\text{id}}, \mathbf{z}_j^{\text{text}}\right)\) and \(\hat{\mathbf{I}}_{j, \mathbf{z}_i^{\text{text}}}=\mathcal{D}\left(\mathbf{v}_j^{\text{dir}}, \mathbf{z}_j^{\text{app}}, \mathbf{z}_j^{\text{id}}, \mathbf{z}_i^{\text{text}}\right)\), the ideal outcome would be that these two images are nearly identical even interchangeable. However, as depicted in Figure \ref{fig:tsne}, their feature distances within the latent space are disordered and unpredictable.

Therefore, to learn a disentangled latent space, we propose an id-consistent loss function (Equation \ref{eq:lossidc}) designed to ensure that images rendered from any viewpoint of the same object exhibit the natural similarities in semantic information, color, texture, and lighting. 
\begin{equation}
\mathcal{L}_{idc} = -\log \frac{\exp(I_i^m \cdot I_i^n / \tau)}{\exp(I_i^m \cdot I_i^n / \tau) + \sum_{\substack{j \neq i}}\exp(I_i^m \cdot I_j / \tau)}
\label{eq:lossidc}
\end{equation}

Where \( D \) represent the entire Objaverse renders dataset, where \( O_i \) denotes the \( i \)-th object. The superscripts \( I_i^1 \), \( I_i^2 \), \( I_i^3 \), and \( I_i^4 \) represent the 512x512 RGB images of the object \( O_i \) rendered from four known viewpoints, while \( I_i^5 \), \( I_i^6 \), \( I_i^7 \), and \( I_i^8 \) represent the corresponding four target views. In each iteration, \( I_i^m \) and \( I_i^n \) (\( \left \{ m,n \right \} \subset \left \{ 1,2,3,4,5,6,7,8 \right \} \)) are sampled from a batch. Our multi-view dataset inherently provides well-structured positive and negative sample pairs, where different views of the same object serve as positive samples, and the same view of different objects serve as negative samples. We observed that the corresponding latent variables exhibit a form of disentanglement, separating the object's material and color information from its structural features. This facilitates the base generative model's understanding of the 3D essence.

When training the generator, the total loss is defined as Equation \ref{eq:loss}. During training discriminator, the loss remains the same as the original AutoencoderKL loss.

\begin{equation}
    \mathcal{L}_{ID} = \lambda * \mathcal{L}_{idc} + \mathcal{L}_{rec} + \mathcal{L}_{kl}
    \label{eq:loss}
\end{equation}

The parameter $\lambda$ balances the magnitude of the id-consistent loss and other losses, and it decreases as the number of training epochs increases.

\begin{figure}[t]
\centering
\includegraphics[width=1\columnwidth]{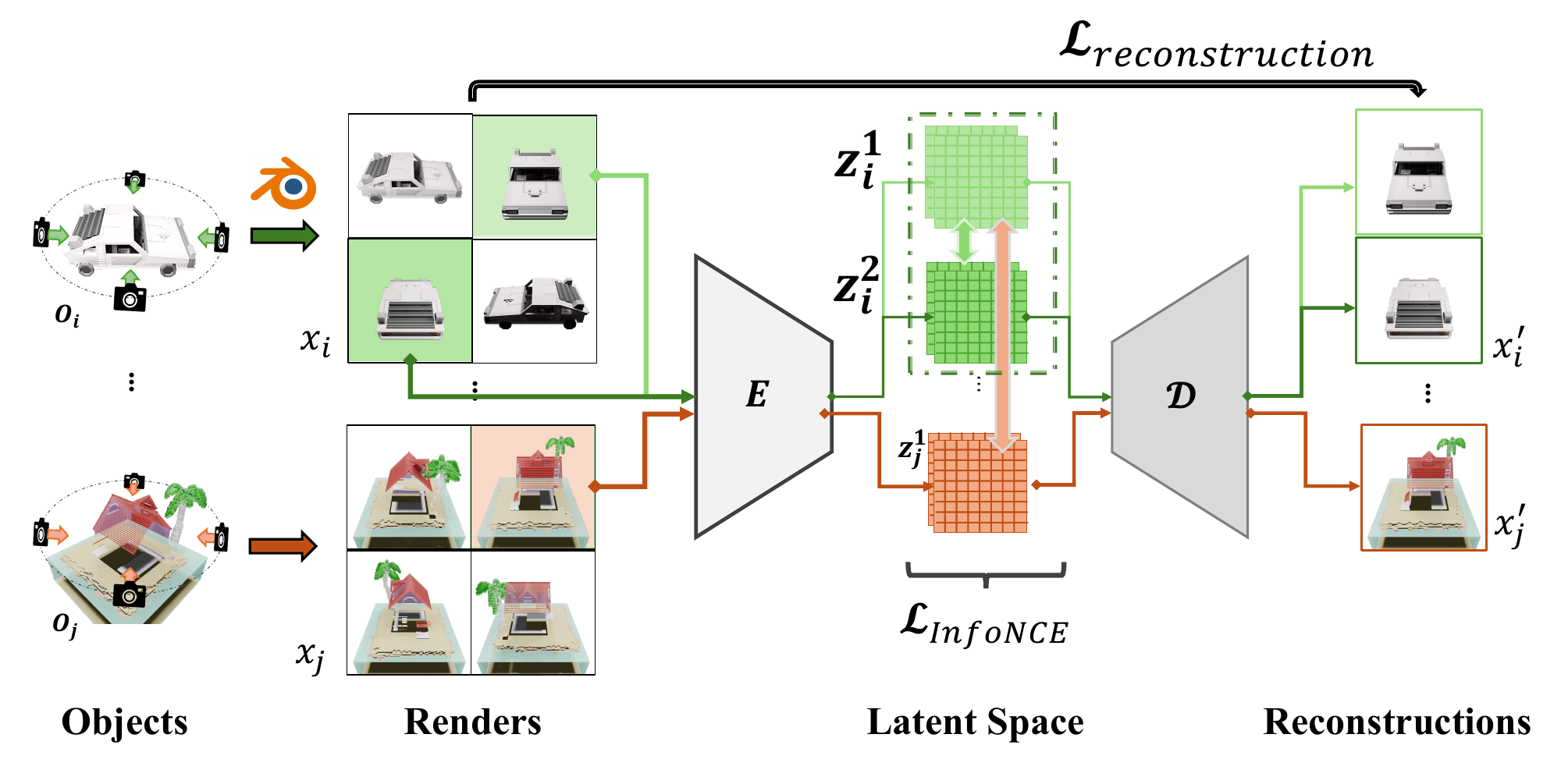} 
\caption{ID-Consistent and Perspective-Disentangled 3D Latent Space Training: By optimizing with our proposed $L_{ID}$, our latent space effectively disentangle identity features from view-dependent characteristics.}
\label{vae}
\end{figure}

\subsection{3.3 Pseudo 3D Multi-View Cross Attention}

As previously mentioned, due to the limitations of single-viewpoint models, earlier 3D reconstruction methods heavily relied on the inductive bias of generative models to infer the unseen regions, which was insufficient for accurate 3D reconstruction. In our few-shot scenario, we utilize sparse viewpoints covering the entire object (specifically, 4 viewpoints) as conditioning for the 2D diffusion model. We fix the four input viewpoints at specific angles and generate four corresponding viewpoints. This approach avoids errors introduced by viewpoint estimation, as seen in previous works like "One-2-3-45", which incorporated dedicated modules for viewpoint estimation, adding extra errors to the process.

\begin{figure}[h]
\centering
\includegraphics[width=1\columnwidth]{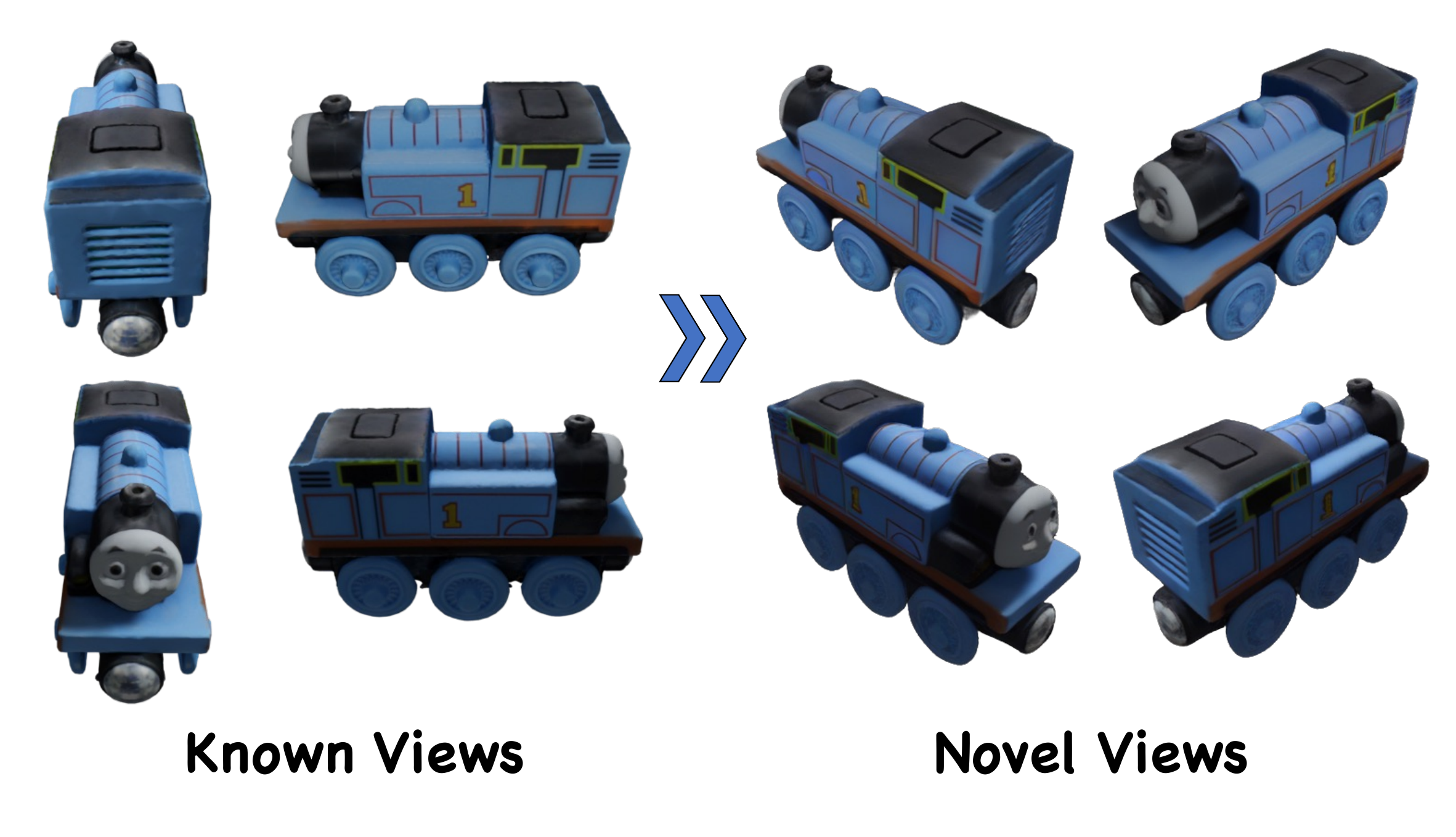} 
\caption{From known viewpoint Conditions to generation of novel perspectives.}
\label{mutil_view}
\end{figure}

As illustrated in Figure~\ref{mutil_view}, we inject multi-view conditions in a four-grid format. This allows the features before self-attention in each layer of the model to be represented in the spatial domain as $Q = (Q_1, Q_2, Q_3, Q_4)$, $K = (K_1, K_2, K_3, K_4)$, and $V = (V_1, V_2, V_3, V_4)$. When these features pass through the model's inherent self-attention layers, the self-attention in the spatial domain evolves into cross-attention among multiple views:
\begin{equation}
\begin{split}
\text{Attn}(Q, K, V) & = \text{softmax}\left(\frac{Q K^T}{\sqrt{d_k}}\right)V \\
& = \begin{pmatrix}
\sum_{j=1}^{4} \text{softmax}\left(\frac{Q_1 K_j^T}{\sqrt{d_k}}\right) V_j \\
\sum_{j=1}^{4} \text{softmax}\left(\frac{Q_2 K_j^T}{\sqrt{d_k}}\right) V_j \\
\sum_{j=1}^{4} \text{softmax}\left(\frac{Q_3 K_j^T}{\sqrt{d_k}}\right) V_j \\
\sum_{j=1}^{4} \text{softmax}\left(\frac{Q_4 K_j^T}{\sqrt{d_k}}\right) V_j
\end{pmatrix}
\end{split}
\end{equation}

Therefore, such condition injection seamlessly introduces cross-attention interactions among different 3D viewpoints without any additional computational cost. 

Essentially, this pseudo cross-attention mechanism, which incurs no additional computational overhead, facilitates the model's deep comprehension of variations in texture, lighting, and other attributes across different viewpoints of a 3D object. It enables the model to learn the natural pixel-level transformations necessary for transitioning from one viewpoint to another. Ultimately, our model can accurately infer the appearance of unseen viewpoints, as opposed to previous approaches that heavily relied on the generative model's imaginative capabilities.

\begin{figure*}[h!]
\centering
\includegraphics[width=1\textwidth]{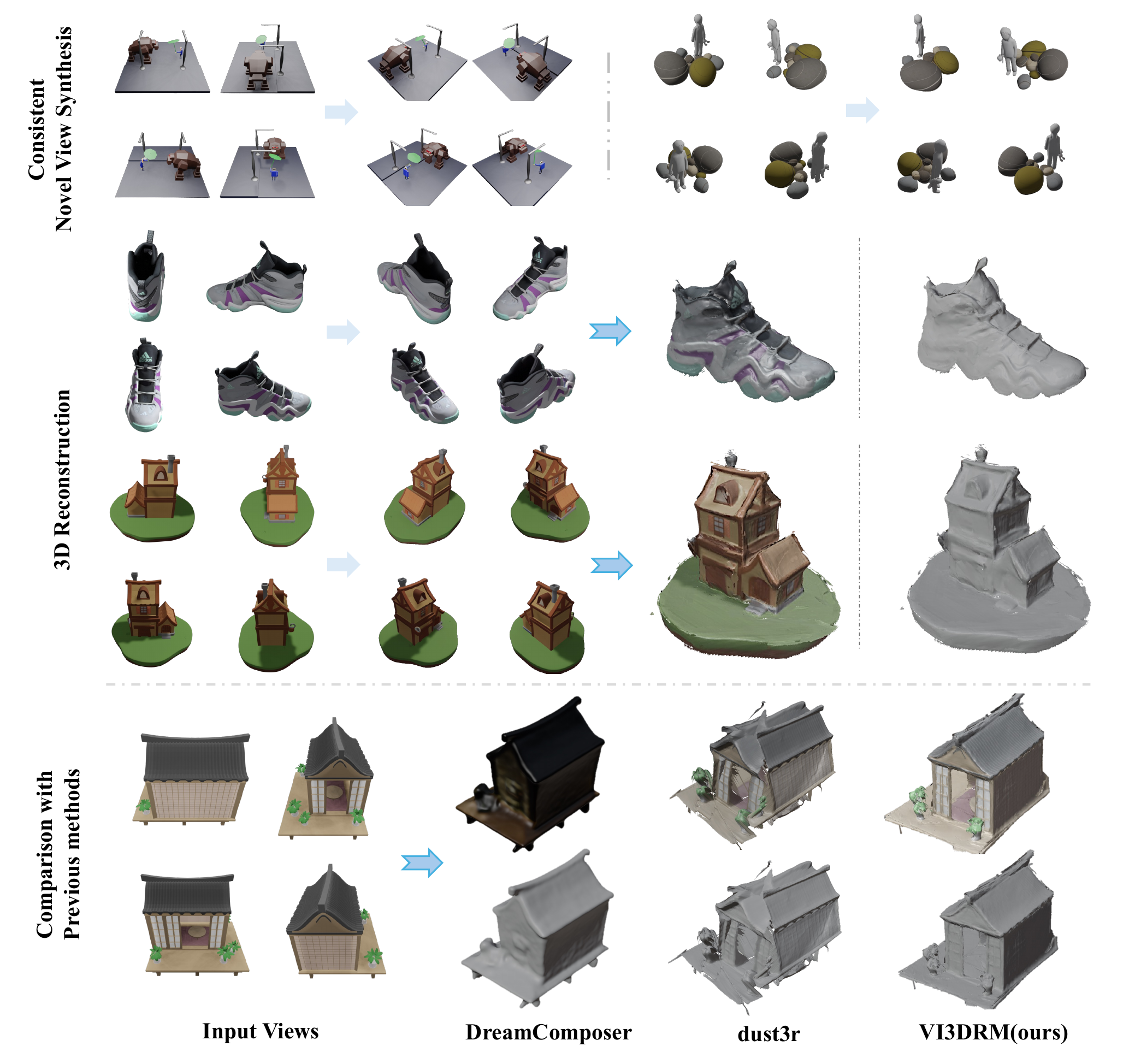} 
\caption{Our method, \textbf{VI3DRM}, can generate realistic novel view images and detailed textured meshes within 60 seconds.}
\label{fig:shows}
\end{figure*}

\subsection{3.4 Diffusion-Based NVS Model}

Given the input perspectives $I_1$, $I_2$, $I_3$, $I_4$, we want to produce $I_5$, $I_6$, $I_7$, $I_8$ using 2d diffusion model. As Section 3.3 discussed, in order to integrate the correlation between the image contents and achieve more consistent model outputs, we combine the conditional images into $I_{con}$, a large image with 1024*1024 resolution and output $I_pred$, the spliced large image(1024*1024 as well) at one time. Formally speaking, we use 2D diffusion as a mapping function $f$ such that $I_{pred} = f(I_{con})$. During training, the condition image $I_{con}$ is first encoded to our expressive and ID-consistent latent $z_{cond}$. In addition to spatial information, we obtained the semantic information $Emb$ of $I_{con}$ through CLIP \cite{radford2021learning} as global condition.

\begin{equation}
    L_{LDM}:=\mathbb{E}_{\mathcal{E}(I_{gt}),I_{con},\epsilon\thicksim\mathcal{N}(0,1),t}\left[\|\epsilon\boldsymbol{-}\epsilon_\theta(z_{gt}^t,t,z_{con},Emb)\|_2^2\right]
\end{equation}

\subsubsection{Reference Attention. }

We adopt Reference Attention \cite{ReferenceAttention} to guide the generation of novel view synthesis using disentangled feature latents from known view images. Specifically, the same U-Net model is applied to the condition latents, and its self-attention key and value matrices are appended to the corresponding attention layers of the model input. 

\subsubsection{Classifier-Free Guidance.}

When training our diffusion model, we adapt Classifier-Free Guidance \cite{ho2022classifier} and apply a dropout rate of 0.3 to the input conditions. During generating novel views, we apply

\begin{equation}
\hat{\mathbf{X}}^0=(1+\sum_{\mathbf{c}\in\mathbf{C}}\lambda_\mathbf{c})\cdot\mathcal{H}(\mathbf{X}^t,t,\mathbf{C})-\sum_{\mathbf{c}\in\mathbf{C}}\lambda_c\cdot\mathcal{H}(\mathbf{X}^t,t,\mathbf{C}|_{\mathbf{c}=\emptyset})
\end{equation}

where $\lambda_\mathbf{c}$ is the CFG scale for condition c. $\mathbf{C}|_{\mathbf{c}=\emptyset}$ denotes that the condition $\mathbf{c}$ is replaced with $\emptyset$.

\subsection{3.5 3D Reconstruction}

Using all these 8 views, we are now able to get accurate 3d reconstruction. We adopt a recently open-sourced algorithm, dust3r \cite{wang2024dust3r}, a dense 3D reconstruction framework to lift images to 3D. Despite dust3r holds the capability of estimating camera pose from un-calibrated images, we observed in our experiments that there remains a certain degree of deviation between the angles estimated by dust3r and the actual angles, which lead to a huge impact on the final reconstruction accuracy. We refer to this tutorial \cite{dust3rpose} and set the angles of images, thereby eliminating the bias introduced by the pose estimation in dust3r. With total 8 views, including both known and prediction images, dust3r first conduct pointmaps for each images pair and perform Global Alignment to get more accurate pointmaps result. 3D representations like accurate point clouds or detailed mesh then can be extracted from the pointmaps.

\section{4. Experiments}
In this chapter, we will present our experimental details and results in detail, including the two main components of our model: the ID-Consistent and Perspective Disentangled 3D Latent Space, and the multi-view conditioned diffusion model. Details regarding the training and evaluation datasets, as well as the metrics we use, are provided in 4.1. Subsection 4.2 elaborates on the training specifics of the our latent space. The training details and results of the diffusion model are presented in 4.3.

\begin{figure}[t]
\centering
\includegraphics[width=1\columnwidth]{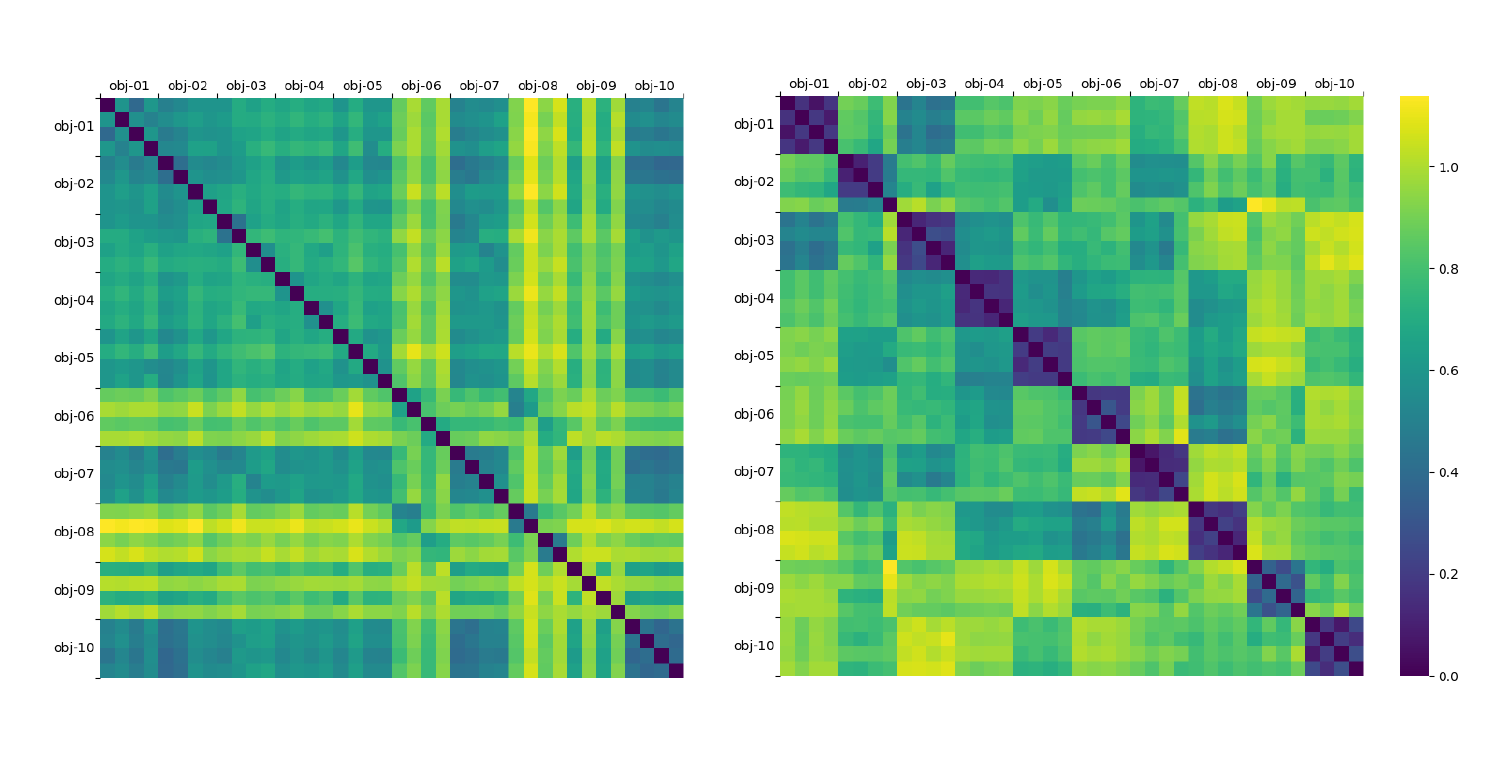}
\caption{This figure illustrates the Cosine Distance Matrices for both the vanilla latent space (left) and our latent space (right). In the right image, the 4x4 darker regions represent the feature similarity of the same object across four different views in our latent space. In contrast, the left image shows that only self-similarity is preserved in the original latent space. Our latent space demonstrates superior decoupling capabilities, as evidenced by the darker areas, indicating a reduced cosine distance and better preservation of the object's identity features compared to the vanilla latent space.}
\label{fig:dist}
\end{figure}

\subsection{4.1 Datasets and Metrics}
\subsubsection{Training Dataset.}
We conducted multi-view rendering on the widely utilized large-scale Objaverse dataset \cite{deitke2023objaverse}. For our experiments, we utilized approximately half of the dataset, encompassing around 400,000 objects. Each object was rendered from eight fixed viewpoints, with an image resolution of 512x512 pixels, against a black background. The lighting was randomized, and the rendering was performed using the EEVEE engine.

\subsubsection{Evaluation Dataset.}
To assess the generalization capability of our model on out-of-distribution data, we extended our evaluation dataset from Objaverse to include the Google Scanned Objects (GSO) dataset, which features high-quality scans of everyday household items. This evaluation setup is consistent with that used for previous work DreamComposer \cite{yang2024dreamcomposer}, encompassing 30 objects that include both common household items and various animal species.

\subsubsection{Metrics.}

In line with previous studies, we employ Peak Signal-to-Noise Ratio (PSNR), Structural Similarity Index (SSIM) \cite{wang2004image}, and Learned Perceptual Image Patch Similarity (LPIPS) \cite{zhang2018unreasonable} as evaluation metrics.

\subsection{4.2. ID Consistent Latent Space}
\subsubsection{Training Details.}

We utilized the aforementioned objaverse renders dataset to optimize our ID-consistent latent space. We down-sampled the data resolution from 512 to 384 when training. To balance the InfoNCE Loss and LPIPSWithDiscriminator Loss, we employed a dynamically adjusted $\lambda$ that uniformly decreased from 1e4 to 1e3. The model was trained for total 10 epochs on 8 L40s with a batch size of 2, using the Adam optimizer and an initial learning rate of 4e-5.

\subsubsection{Analyse of Latent Space.}

 We conducted zero-shot testing on the GSO dataset by randomly sampling 10 objects, with 4 perspective renders for each. These 40 images are encoded into both the vanilla latent space and our latent space. As illustrated in Figure \ref{fig:dist}, due to superior decoupling capabilities, our latent space significantly reduces the cosine distance between views latents of the same object.

\subsection{4.3. Multi-View Conditioned 2D Diffusion Model}

\subsubsection{Training Details.}

Staring from Stable Diffusion v2-1 Model, we trained our model with a conservative learning rate of 1e-5, for approximately 100,000 iterations.

\subsubsection{Results.}

\begin{table}[h]
\centering
\renewcommand{\arraystretch}{1.2}
\resizebox{\columnwidth}{!}{
\begin{tabular}{c|c|c|c|c}

Methods & View(s) &PSNR $\uparrow$ & SSIM $\uparrow$ & LPIPS $\downarrow$ \\
\specialrule{.1em}{.05em}{.05em}
Zero123 & 1 & 21.44 & 0.838 & 0.133 \\ 
Zero123 XL & 1 & 20.11 & -  & 0.113 \\ 
One2345++  & 1 & 22.12 & -  & 0.110 \\
DreamComposer & 4 & 25.58 & 0.888 & 0.086 \\
DreamComposer & 6 & 27.21 & 0.906 & 0.073 \\
VI3DRM(Ours) & 4 & \textbf{38.61($\uparrow42\%$)} & \textbf{0.929($\uparrow2\%$)} & \textbf{0.027($\downarrow63\%$)}

\end{tabular}}
\caption{Comparison with previous methods on NVS. Evaluated on the complete GSO dataset. View(s) represent the number of perspectives used. The average metrics are calculated at their respective target angles.}
\label{table:matrics}
\end{table}

We evaluate our model on NVS task on the whole GSO dataset. Since the previous algorithms generate different target perspectives, our indicators are measured by comparing the prediction results and the actual results of the algorithms at their respective target perspectives. The quantitative results are shown in Table \ref{table:matrics}. The qualitative results, including novel view synthesis and 3d reconstruction results, are shown in Figure \ref{fig:shows}. Our model outperforms previous algorithms in terms of the realism of new perspective generation tasks, including details, lighting, and the quality of 3D reconstruction.

\section{5. Conclusion}
We propose VI3DRM, a Sparse-View based 3D reconstruction model which works in an ID-consistent and perspective-disentangled 3D latent space. VI3DRM is able to generate Photo-Realistic novel views and detailed 3D reconstructions, outperforming previous methods significantly. For now we have restricted the viewpoints to fixed angles. In future work, we aim to optimize the model's tolerance for varying viewpoints, thereby making it suitable for a wider range of everyday scenarios.

\pagebreak

\bibliography{aaai25}

\end{document}